\documentclass[11pt,letterpaper]{article}
\providecommand{\TRPrimaryHex}{CF4E08}
\providecommand{\TRInkHex}{252525}
\usepackage{liveaigc_report}
\definecolor{rankbest}{RGB}{169,209,142}
\definecolor{ranksecond}{RGB}{226,239,218}
\renewcommand{\best}[1]{\cellcolor{rankbest}#1}
\newcommand{\second}[1]{\cellcolor{ranksecond}#1}
\renewcommand{\arraystretch}{1}
\newcommand{\rwparagraph}{\paragraph}
\newcommand{\papertablefont}{\fontsize{10}{12}\selectfont}
\usepackage{threeparttable}
\newcommand{\papermetric}[3]{\shortstack{#1\\$#2$\,{\scriptsize\cite{#3}}}}

\titleformat{\paragraph}[runin]
  {\sffamily\bfseries\color{TRPrimary}}{}{0pt}{}
\AtBeginDocument{%
  \fontsize{10.5pt}{12.5pt}\selectfont
  \setlength{\abovedisplayskip}{8pt plus 1pt minus 2pt}%
  \setlength{\belowdisplayskip}{8pt plus 1pt minus 2pt}}
\titlespacing*{\section}{0pt}{12pt plus 1pt minus 2pt}{6pt}
\titlespacing*{\subsection}{0pt}{9pt plus 1pt minus 1pt}{4pt}
\titlespacing*{\paragraph}{0pt}{5pt plus 1pt minus 1pt}{0.42em}
\titleformat{\section}
  {\large\sffamily\bfseries\color{TRPrimary}}
  {\thesection}{0.55em}{}
  [\vspace{1.5pt}\color{TRRule}\titlerule]
\titleformat{\subsection}
  {\normalsize\sffamily\bfseries\color{TRPrimary}}
  {\thesubsection}{0.5em}{}
\titleformat{\subsubsection}
  {\small\sffamily\bfseries\color{TRPrimaryDark}}
  {\thesubsubsection}{0.45em}{}
\title{TaoMate-FVSR:\\Audio-Guided Facial Video Restoration}
\reportauthors{%
  \mbox{Xinyang Li\textsuperscript{$\dagger$}, Zeran Xu\textsuperscript{$\dagger$}, Jingna Sun, Qijun Gan, Chenwei Zhang, Meiguang Jin\textsuperscript{*}, Junfeng Ma}}
\authornote{\textsuperscript{$\dagger$}Equal contribution.\qquad\textsuperscript{*}Corresponding author.}
\affiliation{Alibaba Group}
\teamname{%
  \mbox{\href{mailto:lyon.lxy@taobao.com}{\texttt{lyon.lxy@taobao.com}}\quad
  \href{mailto:xuzeran.xzr@taobao.com}{\texttt{xuzeran.xzr@taobao.com}}\quad
  \href{mailto:meiguang.jmg@taobao.com}{\texttt{meiguang.jmg@taobao.com}}}}
\shorttitle{TaoMate-FVSR}
\reportlabel{Technical Report}
\reportdate{November 2026}
\leftlogo[height=0.9cm]{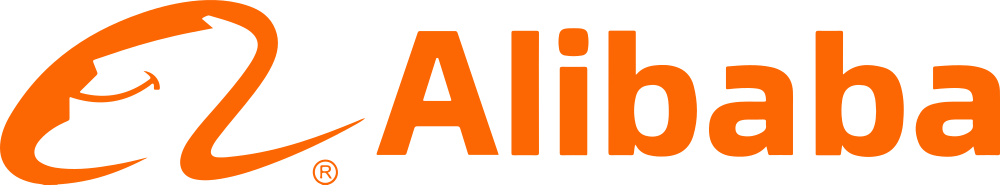}
\rightlogo[height=0.9cm,trim=436bp 638bp 306bp 480bp,clip]{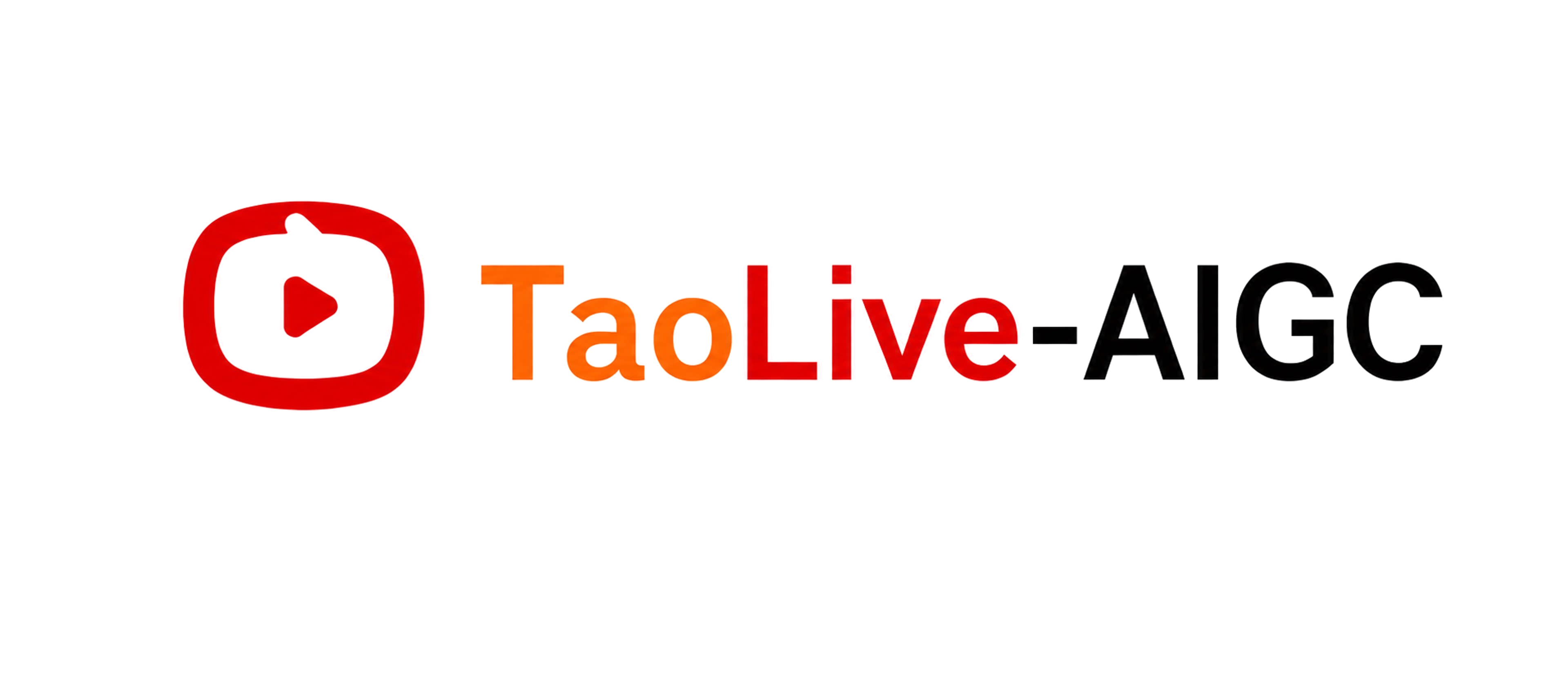}
\projectleaders{}
\corecontributors{}
\contributors{}
\equalcontrib{}
\acknowledgments{}
\hypersetup{
  pdftitle={TaoMate-FVSR: Audio-Guided Facial Video Restoration},
  pdfauthor={Xinyang Li, Zeran Xu, Jingna Sun, Qijun Gan, Chenwei Zhang, Meiguang Jin, Junfeng Ma},
  pdfsubject={Reference-guided facial video restoration and audio-guided visual dubbing},
  pdfkeywords={TaoMate-FVSR, facial video restoration, visual dubbing, reference conditioning, flow matching}
}

\begin{document}
\maketitle
\begin{abstract}
Low-quality facial videos suffer from missing fine details, while AI-generated videos may additionally contain implausible teeth and lip movements that do not match the speech. Restoring the mouth region is particularly challenging: enhancing existing details can preserve structural errors, whereas regenerating them risks changing facial identity and articulation. We propose an audio-guided facial video restoration framework that reconstructs unreliable mouth details and aligns lip movements with the speech while preserving facial identity and temporal consistency. Built on a video diffusion transformer, our framework mixes input video latents with noise to retain useful source information while allowing corrupted details to be regenerated. Audio guides lip movements, reference images constrain facial appearance, and previously generated frames promote temporal consistency. We train the model in two stages, first learning visual restoration from degraded images and videos, then learning audio-guided lip correction from degraded dubbed videos. The resulting model supports both restoration with the original audio and visual dubbing with replacement audio. On talking-head restoration benchmarks, our method achieves the lowest LPIPS and DISTS and the best lip synchronization scores among the evaluated restoration baselines. Qualitative results on AI-generated videos further show improvements in facial detail and mouth-region appearance, including clearer tooth boundaries and lip contours.
\par\smallskip
\noindent\textbf{Project page:} \url{https://taoliveaigc.github.io/TaoMate-FVSR/}
\end{abstract}

\begin{figure*}[!t]
  \centering
  \includegraphics[width=\textwidth]{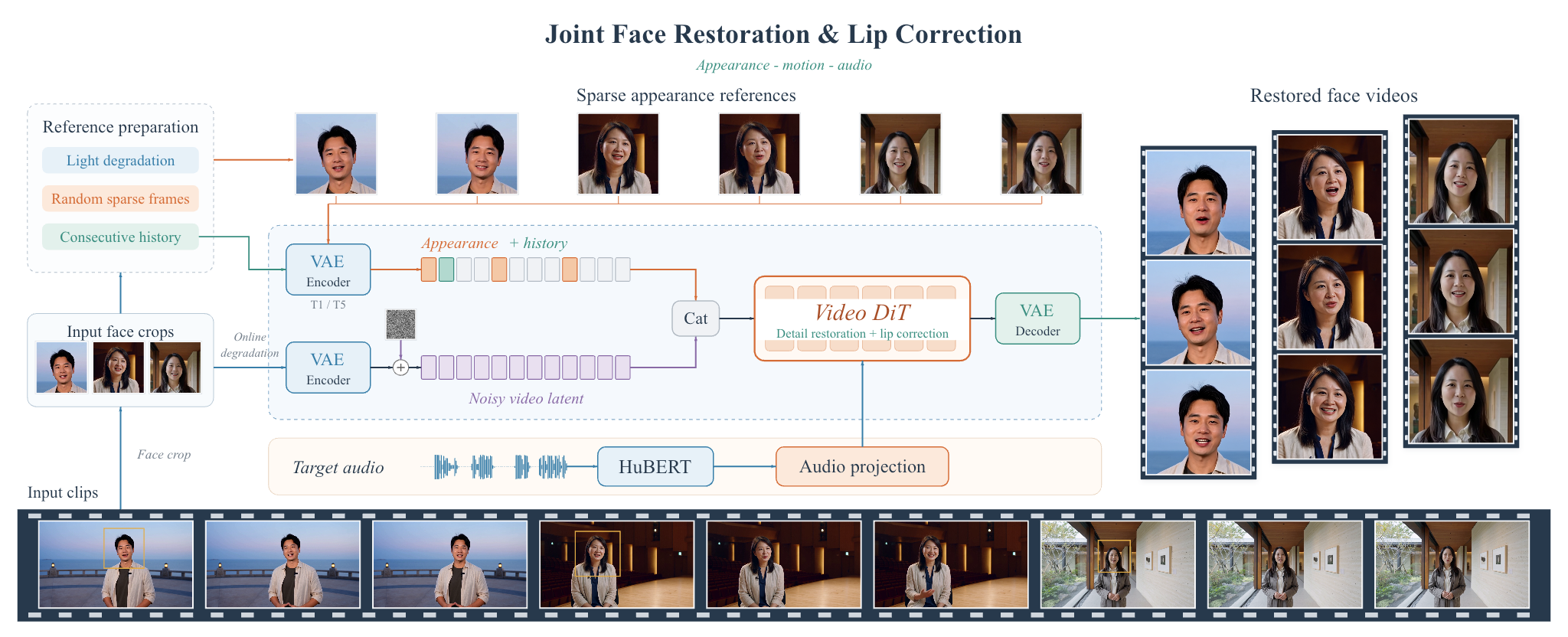}
  \caption{\textbf{Audio-guided facial restoration from AIGC footage.} Small facial regions provide limited evidence for lip and teeth structure. The model restores these regions using reference frames and accompanying audio. The three clips illustrate the source scenes and reconstructed heads; the reference bank depicts training-time conditioning. Quantitative references originate from the GT first frame. ``Cat'' denotes channel-wise fusion implemented by separate linear projections and summation.}
  \label{fig:teaser}
\end{figure*}

\section{Introduction}
\label{sec:intro}
Low-quality facial videos often lack fine details due to limited resolution, blur, and compression. AI-generated videos pose an additional challenge. Video generation models can synthesize detailed scenes and coordinated character motion~\cite{wan2025}, yet faces may remain poorly resolved when they occupy only a small region of the frame. Beyond missing detail, generated faces may contain implausible teeth or lip movements that do not match the speech. These defects become particularly noticeable during close viewing and reframing. Improving facial quality without regenerating the entire shot is therefore valuable for both captured and generated videos.

Teeth are especially difficult to restore because their shape and boundaries must remain consistent as the lips, jaw, and head move. Low resolution can make individual teeth indistinguishable, while generation can introduce incorrect or temporally unstable structures. Sharpening these observations can accentuate their errors; restoration must instead resolve the ambiguous structure across frames.

Recent video super-resolution methods offer powerful generative priors and reference guidance for restoring visual detail. SeedVR2~\cite{wang2025seedvr2} uses generative priors for reconstruction, FlashVSR~\cite{zhuang2025flashvsr} targets efficient streaming restoration, and SparkVSR~\cite{yu2026sparkvsr} introduces sparse high-quality keyframes for appearance control. For speaking faces, however, visual quality alone is insufficient: a sharp and plausible mouth can still move out of sync with the speech. Audio provides complementary cues for recovering these movements. Extreme-scale talking-face upsampling~\cite{hegde2022extreme} combines audio with a high-resolution identity image; MDVD-Net~\cite{zhang2023mdvd} exploits multimodal cues for compressed face video; and GAVN~\cite{cao2026gavn} models temporal and identity information. Building on these directions, we study facial video restoration when the input mouth details may be missing or incorrect, requiring visual reconstruction and audio-guided lip correction within the same framework.

A central challenge is deciding how much of the input to preserve. Strong source conditioning retains appearance and motion but can also preserve mouth errors; greater freedom to regenerate details risks identity drift. Our video diffusion transformer addresses this trade-off by starting reconstruction from noise-mixed input latents. Target speech constrains articulation, while appearance references and recent generated frames supply identity and temporal cues.

Training first learns restoration from degraded images and videos, then introduces dubbed sources whose mouth movements differ from the target. This mismatch trains the model to follow the supplied speech rather than copy source articulation. Independently encoded reference images and sequence-encoded history occupy separate temporal slots, retaining a fixed appearance anchor across video segments. Figure~\ref{fig:teaser} illustrates the task, and Figure~\ref{fig:method} presents the framework.

We evaluate original-audio restoration and replacement-audio dubbing using visual-quality and synchronization metrics, including comparisons with super-resolution and dubbing cascades~\cite{he2026xdub,cheng2022videoretalking}. Native AI-generated examples further examine tooth boundaries, lip contours, and surrounding facial texture.

The contributions are threefold:
\begin{itemize}
    \item We introduce a joint facial reconstruction and speech-guided lip-correction framework for original-audio restoration and replacement-audio dubbing.
    \item We combine source--noise reconstruction with two-stage restoration and lip-correction training, enabling the model to revise unreliable mouth details.
    \item We design structured conditioning with separate appearance and history slots to preserve identity and continuity across video segments.
\end{itemize}

\Needspace{4\baselineskip}
\section{Related Work}
\label{sec:related}
\rwparagraph{Facial and video restoration.}
Facial restoration methods use learned priors to recover details lost through degradation. GAN-based approaches~\cite{wang2021gfpgan,yang2021gpen} and codebook-based approaches~\cite{zhou2022codeformer} provide high-quality facial representations, while diffusion-based methods~\cite{wang2024stablesr,lin2024diffbir,yu2024supir} exploit generative priors for perceptually realistic reconstruction. Video restoration further incorporates temporal information through feature propagation~\cite{chan2022basicvsrpp,feng2024keep} or video diffusion models~\cite{zhou2024upscaleavideo,wang2025seedvr2,zhuang2025flashvsr}. Reference-guided methods recover identity-specific appearance: InstantRestore~\cite{zhang2025instantrestore} uses shared-image attention, RG-FVR~\cite{eteke2026rgfvrpaper} introduces identity conditioning into a flow-based video generator, and SparkVSR~\cite{yu2026sparkvsr} propagates sparse high-quality keyframes. Our task additionally requires correcting unreliable source articulation according to target speech.

\rwparagraph{Audio-assisted facial restoration.}
Audio helps recover mouth movements when visual observations are insufficient. Extreme-scale talking-face upsampling~\cite{hegde2022extreme} combines low-resolution video, audio, and an identity image. MDVD-Net~\cite{zhang2023mdvd} uses audio, geometry, and codec cues; GAVN~\cite{cao2026gavn} models complementary temporal and identity information. We extend this audiovisual restoration setting to degraded dubbed sources whose articulation differs from the target speech.

\rwparagraph{Teeth restoration and temporal consistency.}
HDTR-Net~\cite{li2023hdtr} enhances teeth and surrounding mouth details through fine-grained feature fusion and reference guidance. OrthoNet~\cite{su2025orthonet} addresses teeth artifacts using detail-oriented alignment with long-term and short-term memory. We address dental detail within full-face reconstruction, where its temporal evolution must also follow speech, without a separate teeth-restoration module.

\rwparagraph{Visual dubbing.}
Visual dubbing modifies an existing video's mouth movements to match target speech. Wav2Lip~\cite{prajwal2020wav2lip} uses expert synchronization supervision, while VideoReTalking~\cite{cheng2022videoretalking} combines lip editing with identity-aware enhancement. KeySync~\cite{bigata2025keysync} uses a two-stage framework and masking to address temporal consistency, expression leakage, and occlusions. X-Dub~\cite{he2026xdub} develops mask-free dubbing, and TBDub~\cite{li2026tbdub} extends it through task-adaptive post-training and few-step distillation. LipForcing~\cite{cho2026lipforcing} explores streaming lip synchronization through few-step autoregressive diffusion. In our setting, the model must additionally recover degraded facial detail during dubbing.

\Needspace{6\baselineskip}
\section{Method}
\label{sec:method}
\begin{figure*}[!t]
  \centering
  \includegraphics[width=\textwidth]{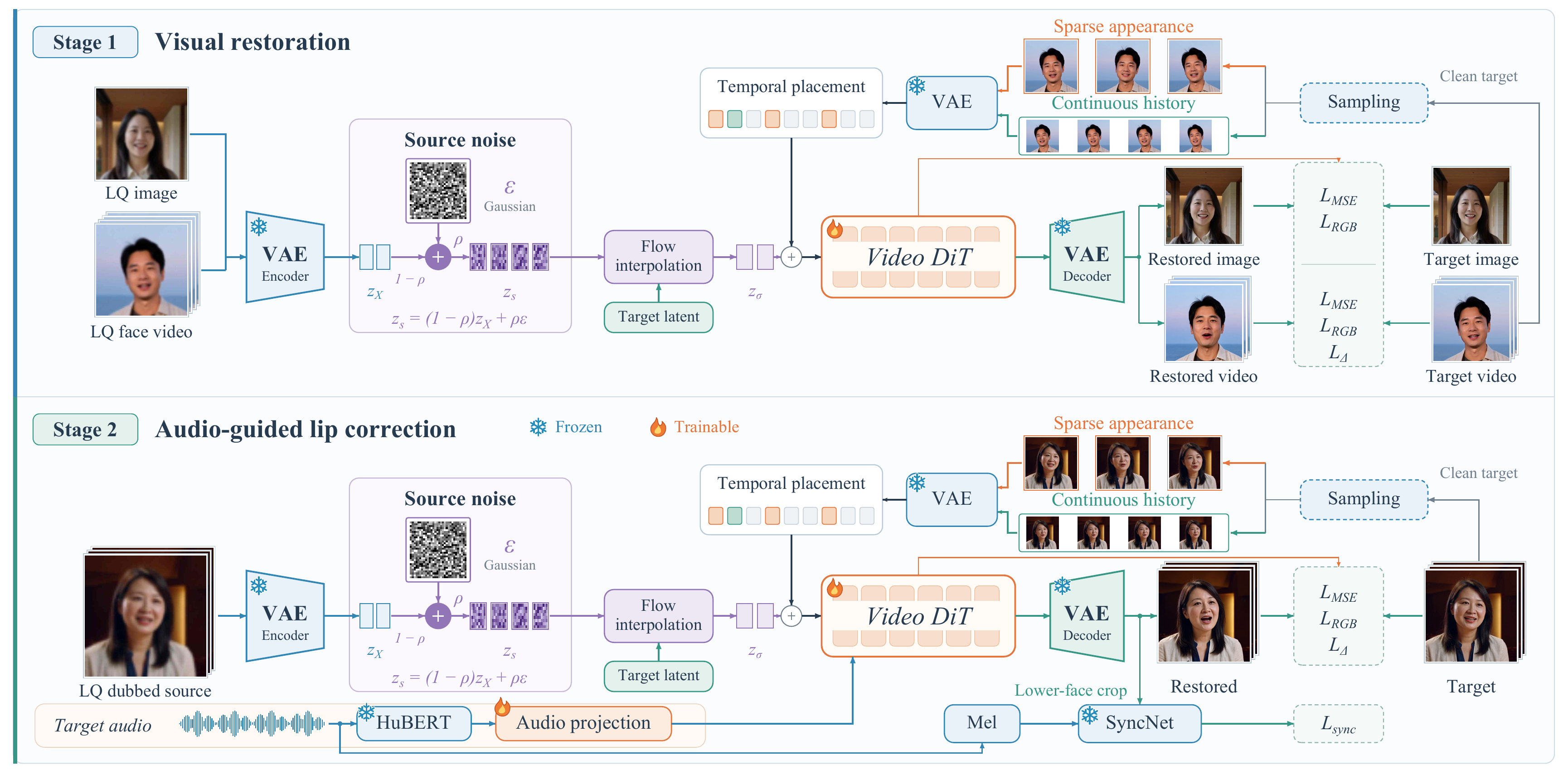}
  \caption{\textbf{Audio-guided facial video restoration.} Stage 1 jointly trains on degraded images and videos with their corresponding targets. Images use only flow-matching MSE ($\mathcal L_{\mathrm{MSE}}$) and RGB losses; videos additionally use temporal supervision. Stage 2 uses degraded dubbed sources and target audio for lip correction. Sparse references and continuous history occupy distinct conditioning slots. The VAE, HuBERT, and SyncNet are frozen; the video DiT and conditioning projections are trainable. AIGC thumbnails illustrate the input and target roles.}
  \label{fig:method}
\end{figure*}

Given a low-quality facial video $X$, target audio $a$, and appearance references $R$, TaoMate-FVSR reconstructs a high-quality video with speech-aligned lip movements. Figure~\ref{fig:method} shows the video diffusion transformer operating in the latent space of a frozen causal video autoencoder. A single trained model supports both original-audio restoration and replacement-audio dubbing.

\subsection{Source--Noise Reconstruction}
\label{sec:flow}
A degraded video provides useful information about facial appearance and motion, but its mouth details may be blurred or incorrect. To retain this information without requiring the output to reproduce unreliable details, we initialize reconstruction from a noisy version of the input latents. A frozen causal video encoder $E$ maps the input $X$ and clean training target $Y$ to $z_X=E(X)$ and $z_Y=E(Y)$. Inputs are resized to the target resolution before encoding. Images are treated as single-frame examples in the same latent representation.

We mix the input latents with Gaussian noise to obtain the reconstruction starting point:
\begin{equation}
z_s=(1-\rho)z_X+\rho\epsilon,\qquad \epsilon\sim\mathcal N(0,I).
\end{equation}
The mixing strength $\rho$ controls how much source information remains in $z_s$. During training, we construct a linear path between the clean target and this noisy source:
\begin{equation}
z_\sigma=(1-\sigma)z_Y+\sigma z_s,\qquad v^\star=z_s-z_Y.
\end{equation}
Following flow matching~\cite{lipman2022flowmatching}, the video DiT predicts the target velocity $v^\star$ from $z_\sigma$, flow time $\sigma$, and the conditioning inputs. The two scalars serve different roles: $\rho$ determines the reconstruction starting point, whereas $\sigma$ specifies a position along the training path. At inference, we integrate the predicted velocity field from $\sigma=1$ to $\sigma=0$, starting at $z_s$.

\subsection{Audio and Reference Conditioning}
\label{sec:history}
\paragraph{Audio conditioning.}
We extract target-speech features with a frozen HuBERT encoder~\cite{hsu2021hubert} and pass them through a trainable audio projection. The projected features are aligned with the reconstruction windows and supplied to the DiT. The same interface accepts original audio or replacement speech.

\paragraph{Appearance and history conditioning.}
Reference images and recent frames provide different information. Reference images supply persistent appearance cues, but their pose and mouth configuration need not match the current output frame. Recent frames provide local motion and appearance context. We therefore encode sparse reference images independently and obtain history features through sequence encoding. Their features occupy separate temporal positions in a reference canvas, with unused positions set to zero. A shared projection adds this canvas to the DiT input features without replacing the reconstruction latents. This design makes appearance references available as conditions rather than prescribing their mouth configuration as the output.
The fixed appearance reference remains accessible in every inference window, independently of generated history.

\subsection{Two-Stage Training}
\label{sec:training}
\paragraph{Stage 1: visual restoration.}
We first train the model using high-quality facial images and videos $Y$ paired with degraded inputs $X=\mathcal D(Y)$. The degraded inputs are resized back to the target resolution before encoding. This stage learns to reconstruct facial appearance from incomplete visual observations. Image examples supervise spatial detail, while video examples additionally supervise changes over time. Sparse clean target frames provide appearance references during training.

For video examples, the reconstruction objective combines latent flow supervision with losses on decoded predictions:
\begin{equation}
\mathcal L_{\mathrm{rec}}=\mathcal L_{\mathrm{flow}}
+0.1\mathcal L_{\mathrm{RGB}}
+0.1\mathcal L_{\Delta}
+0.1\mathcal L_{\mathrm{LPIPS}}.
\end{equation}
The flow loss is a time-weighted velocity MSE, denoted $\mathcal L_{\mathrm{MSE}}$ in Figure~\ref{fig:method}, with twice the weight assigned to the mouth region. RGB MSE and LPIPS~\cite{zhang2018lpips} supervise reconstruction fidelity and perceptual similarity, respectively. The temporal term $\mathcal L_\Delta$ matches adjacent-frame differences between predictions and targets. Still images use only $\mathcal L_{\mathrm{flow}}+0.1\mathcal L_{\mathrm{RGB}}$, without perceptual or temporal losses. To compute decoded video losses, we sample a four-latent window and use detached preceding predictions as causal decoding context.

\paragraph{Stage 2: audio-guided lip correction.}
Paired visual restoration does not explicitly require the model to change source articulation. We therefore continue training with dubbed sources $X^{\mathrm{dub}}$ and clean targets $Y$ that share the person and scene but differ in articulation. We degrade the dubbed source to obtain $X=\mathcal D(X^{\mathrm{dub}})$ and use the audio associated with $Y$ as the target condition. Reconstructing $Y$ now requires both visual detail recovery and lip correction, discouraging the model from copying the source mouth.

We retain the reconstruction losses and add audiovisual synchronization supervision:
\begin{equation}
\mathcal L_{\mathrm{stage2}}=\mathcal L_{\mathrm{rec}}
+\lambda_{\mathrm{sync}}\mathcal L_{\mathrm{sync}},\qquad \lambda_{\mathrm{sync}}=10^{-4}.
\end{equation}
A frozen SyncNet provides the synchronization loss, following Wav2Lip~\cite{prajwal2020wav2lip}. The VAE, HuBERT, and SyncNet remain frozen, while the video DiT and conditioning projections are trainable. Training uses 77-frame windows corresponding to 20 temporal latents, with source--noise strength $\rho\in[0.80,0.95]$. Appearance and history conditions are independently dropped with probabilities 0.1 and 0.4; the second stage additionally drops audio with probability 0.1.

\subsection{Long-Video Inference}
\label{sec:inference}
\paragraph{History across windows.}
We reconstruct long videos in overlapping windows while keeping a fixed appearance reference $r$. For window $k$, let $\tau$ denote the index of the last retained output frame from the preceding window. We construct the conditioning sequence
\begin{equation}
P_k=\bigl[r,\hat y_{\tau-3},\hat y_{\tau-2},\hat y_{\tau-1},\hat y_{\tau}\bigr].
\end{equation}
The reference $r$ is encoded independently, while the second temporal latent of $E(P_k)$ supplies the history feature. These features are placed in their respective conditioning slots. Thus, the history feature is obtained by encoding the reference-prefixed sequence, rather than encoding the four history frames independently. For the initial window, we repeat $r$ five times because no generated history is available. Later windows use previously reconstructed frames without access to future targets.

\paragraph{Window transitions.}
We additionally encode the last retained frame as $E(\hat y_\tau)$ to provide an appearance condition at the transition to new content. This independent encoding complements the sequence-encoded history feature. Within the overlap, we reuse the preceding window's latent trajectory at matching flow times. Source frames, noise, and audio conditions remain aligned to the same video timeline.

\paragraph{Stateful decoding.}
The causal VAE decoder preserves its internal state when decoding newly retained latents from successive windows. Each window contains five conditioning frames and 72 payload frames. A stride of 56 yields an overlap of 16 payload frames. We discard the conditioning prefix and duplicated overlap to recover the original video duration. Outputs are assembled without RGB cross-fading or histogram matching.

\FloatBarrier
\section{Experiments}
\label{sec:experiments}
We compare original-audio restoration and replacement-audio dubbing quantitatively, examine native AI-generated videos qualitatively, and ablate the reconstruction and conditioning designs.
\begin{table*}[!t]
\centering
\papertablefont
\setlength{\tabcolsep}{2.8pt}
\renewcommand{\arraystretch}{1}
\arrayrulecolor{black!30}
\caption{\textbf{Facial restoration under original audio.} Results on the 150-video benchmark, averaged equally over common valid clips. Darker and lighter green mark the best and second-best results, respectively.}
\label{tab:restoration150}
\begin{threeparttable}
\begin{tabular}{@{}lrrrrrrrrrr@{}}
\toprule
& \multicolumn{4}{c}{Reference-based fidelity} & \multicolumn{2}{c}{Lip sync.} & \multicolumn{4}{c}{No-reference quality} \\
\cmidrule(lr){2-5}\cmidrule(lr){6-7}\cmidrule(l){8-11}
\shortstack[l]{Method\\\strut} & \multicolumn{1}{c}{\papermetric{PSNR}{\uparrow}{wang2009mse}} & \multicolumn{1}{c}{\papermetric{SSIM}{\uparrow}{wang2004ssim}} & \multicolumn{1}{c}{\papermetric{LPIPS}{\downarrow}{zhang2018lpips}} & \multicolumn{1}{c}{\papermetric{DISTS}{\downarrow}{ding2022dists}} & \multicolumn{1}{c}{\papermetric{Sync-C}{\uparrow}{chung2016syncnet}} & \multicolumn{1}{c}{\papermetric{Sync-D}{\downarrow}{chung2016syncnet}} & \multicolumn{1}{c}{\papermetric{MUSIQ}{\uparrow}{ke2021musiq}} & \multicolumn{1}{c}{\papermetric{CLIP-IQA}{\uparrow}{wang2023clipiqa}} & \multicolumn{1}{c}{\papermetric{FasterVQA}{\uparrow}{wu2022fasterquality}} & \multicolumn{1}{c}{\papermetric{DOVER}{\uparrow}{wu2023dover}} \\
\midrule
GT & -- & -- & -- & -- & 6.959 & 7.863 & 62.70 & 0.4525 & 0.7735 & 0.6148 \\
\midrule
GPEN & 26.77 & 0.7970 & 0.2267 & 0.1825 & 6.309 & 8.288 & \second{76.47} & \second{0.7032} & \second{0.8931} & \second{0.8505} \\
CodeFormer & 27.38 & 0.8045 & 0.2195 & 0.1876 & 5.856 & 8.693 & 74.42 & 0.6649 & 0.8854 & 0.7954 \\
InstantRestore & 27.42 & 0.8283 & \second{0.1380} & \second{0.1421} & 6.209 & 8.403 & 68.11 & 0.5166 & 0.8562 & 0.7704 \\
SparkVSR & \best{30.62} & \best{0.8863} & 0.1755 & 0.1586 & 6.545 & 8.170 & 65.66 & 0.4417 & 0.8459 & 0.7467 \\
RG-FVR & 26.69 & 0.8308 & 0.1645 & 0.1461 & 5.685 & 9.024 & 70.51 & 0.5246 & 0.8688 & 0.7407 \\
SeedVR2-3B & 27.49 & 0.8456 & 0.2227 & 0.1630 & 6.377 & 8.339 & 57.36 & 0.4412 & 0.7548 & 0.5601 \\
SeedVR2-7B & 28.60 & 0.8584 & 0.2046 & 0.1525 & 6.359 & 8.319 & 54.30 & 0.3939 & 0.7433 & 0.5212 \\
FlashVSR-Tiny & 27.74 & 0.8497 & 0.1563 & 0.1428 & 6.490 & 8.275 & 70.19 & 0.5395 & 0.8810 & 0.7709 \\
FlashVSR-Full & 27.52 & 0.8435 & 0.1652 & 0.1458 & 6.468 & 8.316 & 71.23 & 0.5776 & 0.8820 & 0.7729 \\
FlashVSR+ & 26.68 & 0.8379 & 0.1652 & 0.1473 & 6.524 & 8.230 & 69.83 & 0.4947 & 0.8775 & 0.7851 \\
\midrule
Ours & \second{28.67} & \second{0.8836} & \best{0.0793} & \best{0.0858} & \second{7.413} & \second{7.561} & 66.84 & 0.4908 & 0.8245 & 0.7178 \\
Ours$^{+}$ & 23.52 & 0.7525 & 0.2402 & 0.1902 & \best{7.453} & \best{7.520} & \best{77.62} & \best{0.7081} & \best{0.8997} & \best{0.8823} \\
\bottomrule
\end{tabular}
\begin{tablenotes}[flushleft]
\footnotesize
\item[] Reference-capable baselines and Ours use the same GT first frame; $^{+}$ denotes GPT-Image-assisted enhancement of that frame for Ours$^{+}$. Quantitative results use fixed-reference motion conditioning. GT uses undegraded source frames and original audio; paired self-comparisons are omitted, and GT is excluded from rankings. Scores use complete prepared $512\times512$ frames (Section~\ref{sec:restoration150}).
\end{tablenotes}
\end{threeparttable}
\end{table*}

\subsection{Experimental Setup}
\paragraph{Evaluation data.}
The original-audio benchmark contains 150 talking-head videos from MEAD~\cite{wang2020mead}, HDTF~\cite{zhang2021hdtf}, TalkVid~\cite{chen2025talkvid}, and our internal dataset. Inputs are synthetically degraded using the training-style degradation pipeline, including $4\times$ downsampling followed by resizing to $512\times512$. The undegraded videos serve as paired targets. Replacement-audio evaluation uses 30 clips and their corresponding replacement-audio pairs, for which no paired target-mouth video is available. We additionally examine four native MiniMax-H3~\cite{minimax2026h3} text-to-audio-video clips without synthetic degradation or clean targets. Ablations use 30 videos, comprising 15 MEAD and 15 HDTF clips.

\paragraph{Reference protocol.}
For quantitative restoration in Table~\ref{tab:restoration150}, reference-capable baselines and Ours use the same undegraded GT first frame. Ours$^{+}$ uses a GPT-Image-enhanced version of that frame, with the model and sampler unchanged. For replacement-audio evaluation, SparkVSR and Ours$^{+}$ share the same enhanced GT first frame. Methods without a reference interface retain their supported inputs. For native AIGC clips, references are enhanced from the input first frame and shared across reference-capable methods. All ablations use the unenhanced GT first frame.

\paragraph{Baselines.}
Restoration comparisons include face restoration methods GPEN~\cite{yang2021gpen} and CodeFormer~\cite{zhou2022codeformer}; reference-guided methods InstantRestore~\cite{zhang2025instantrestore}, SparkVSR~\cite{yu2026sparkvsr}, and RG-FVR~\cite{eteke2026rgfvrpaper}; and video restoration methods SeedVR2~\cite{wang2025seedvr2} and three FlashVSR~\cite{zhuang2025flashvsr} variants. For replacement audio, we compare against ten cascades formed by FlashVSR Full or SparkVSR followed by X-Dub~\cite{he2026xdub}, KeySync~\cite{bigata2025keysync}, TBDub Teacher or Student~\cite{li2026tbdub}, or LipForcing~\cite{cho2026lipforcing}. Baselines retain their released interfaces and sampling budgets.

\paragraph{Metrics and implementation.}
PSNR~\cite{wang2009mse}, SSIM~\cite{wang2004ssim}, LPIPS~\cite{zhang2018lpips}, and DISTS~\cite{ding2022dists} assess fidelity and perceptual similarity when paired targets are available. Sync-C and Sync-D measure audiovisual agreement using frozen SyncNet-v2~\cite{chung2016syncnet}. MUSIQ~\cite{ke2021musiq} and CLIP-IQA~\cite{wang2023clipiqa} assess no-reference frame quality, while FasterVQA~\cite{wu2022fasterquality} and DOVER~\cite{wu2023dover} assess overall video quality. We inspect tooth boundaries and lip contours qualitatively in matched-time crops. Quantitative results are averaged equally over common valid videos using native uncompressed RGB outputs; visual metrics use complete prepared frames. Our quantitative runs use seed 42, source--noise strength 0.8, ten Euler steps, shift 5, audio guidance 4, and reference guidance 1.5.

% Generated from measured comparison.json; see results/crossaudio_cascades30/table_receipt.json.
\begin{table*}[!t]
\centering
\papertablefont
\setlength{\tabcolsep}{5pt}
\renewcommand{\arraystretch}{1}
\arrayrulecolor{black!30}
\caption{\textbf{Visual dubbing under replacement audio.} Equal-video means on the same 30 clips and replacement-audio pairs. Darker and lighter green mark the best and second-best method results.}
\label{tab:crossaudio-cascades30}
\begin{threeparttable}
\begin{tabular}{@{}lrrrrrr@{}}
\toprule
& \multicolumn{2}{c}{Lip sync.} & \multicolumn{4}{c}{No-reference quality} \\
\cmidrule(lr){2-3}\cmidrule(l){4-7}
\shortstack[l]{Method\\\strut} & \multicolumn{1}{c}{\papermetric{Sync-C}{\uparrow}{chung2016syncnet}} & \multicolumn{1}{c}{\papermetric{Sync-D}{\downarrow}{chung2016syncnet}} & \multicolumn{1}{c}{\papermetric{MUSIQ}{\uparrow}{ke2021musiq}} & \multicolumn{1}{c}{\papermetric{CLIP-IQA}{\uparrow}{wang2023clipiqa}} & \multicolumn{1}{c}{\papermetric{FasterVQA}{\uparrow}{wu2022fasterquality}} & \multicolumn{1}{c}{\papermetric{DOVER}{\uparrow}{wu2023dover}} \\
\midrule
\multicolumn{7}{l}{\textit{FlashVSR Full $\to$ dubbing}} \\
\quad X-Dub & 6.137 & 9.332 & 73.25 & \second{0.6582} & \second{0.8750} & \second{0.7845} \\
\quad KeySync & 7.347 & 7.840 & 68.07 & 0.4908 & 0.8736 & 0.7647 \\
\quad TBDub Teacher & 7.736 & 7.543 & 70.18 & 0.6208 & 0.8650 & 0.7179 \\
\quad TBDub Student & 7.469 & 8.196 & 69.72 & 0.5834 & 0.8641 & 0.7309 \\
\quad LipForcing & \second{7.917} & \second{7.186} & 66.18 & 0.4973 & 0.8476 & 0.6958 \\
\addlinespace[3pt]
\multicolumn{7}{l}{\textit{SparkVSR $\to$ dubbing}} \\
\quad X-Dub & 5.853 & 9.409 & \second{74.12} & 0.6335 & 0.8596 & 0.7822 \\
\quad KeySync & 7.175 & 7.964 & 62.17 & 0.4271 & 0.8318 & 0.7199 \\
\quad TBDub Teacher & 7.208 & 7.817 & 67.94 & 0.5383 & 0.8389 & 0.7030 \\
\quad TBDub Student & 7.090 & 8.401 & 67.25 & 0.4987 & 0.8412 & 0.7174 \\
\quad LipForcing & 7.858 & 7.224 & 60.70 & 0.4270 & 0.8079 & 0.6517 \\
\midrule
Ours$^{+}$ & \best{8.175} & \best{7.083} & \best{77.56} & \best{0.7070} & \best{0.8958} & \best{0.8709} \\
\bottomrule
\end{tabular}
\begin{tablenotes}[flushleft]
\footnotesize
\item[] $^{+}$ denotes GPT-Image-assisted enhancement of the GT first frame. SparkVSR and Ours$^{+}$ use the same enhanced reference. Arrows indicate execution order. Motion conditioning follows Table~\ref{tab:restoration150}.
\end{tablenotes}
\end{threeparttable}
\end{table*}

\subsection{Facial Restoration under Original Audio}
\label{sec:restoration150}
\paragraph{Reconstruction and synchronization.}
Table~\ref{tab:restoration150} shows that Ours achieves the lowest paired perceptual errors, with LPIPS of 0.0793 and DISTS of 0.0858. Its Sync-C/D of 7.413/7.561 also improves on SparkVSR at 6.545/8.170, the strongest restoration baseline on these synchronization measures. SparkVSR achieves the highest PSNR and SSIM, exceeding Ours by 1.95\,dB and 0.0027, respectively. Ours therefore leads in perceptual similarity and synchronization, while SparkVSR retains an advantage in pixel fidelity.

\paragraph{Effect of reference enhancement.}
Ours$^{+}$ achieves the highest scores on all four no-reference quality metrics, including MUSIQ of 77.62 and DOVER of 0.8823, while retaining Sync-C/D of 7.453/7.520. Compared with Ours, PSNR decreases from 28.67 to 23.52 and LPIPS increases from 0.0793 to 0.2402. Reference enhancement thus yields higher no-reference quality scores but lower paired fidelity to the original target.

\subsection{Visual Dubbing under Replacement Audio}
\label{sec:crossaudio-cascades30}
Table~\ref{tab:crossaudio-cascades30} compares joint reconstruction with restoration--dubbing cascades. All methods receive the same degraded source and replacement audio, with no additional degradation between cascade stages.

Ours$^{+}$ achieves the best scores across all six reported metrics among the evaluated pipelines. Its Sync-C/D of 8.175/7.083 improves on the strongest synchronization cascade, FlashVSR Full followed by LipForcing, at 7.917/7.186. MUSIQ reaches 77.56, compared with 74.12 for the best cascade on this metric, SparkVSR followed by X-Dub. Different cascades favor different criteria: LipForcing gives the strongest synchronization results, whereas X-Dub leads several quality measures. These results support joint reconstruction as an effective alternative to the evaluated restoration--dubbing cascades under replacement audio.

\subsection{Native AIGC Results}
\begin{figure*}[tbp]
  \centering
  \includegraphics[width=\textwidth]{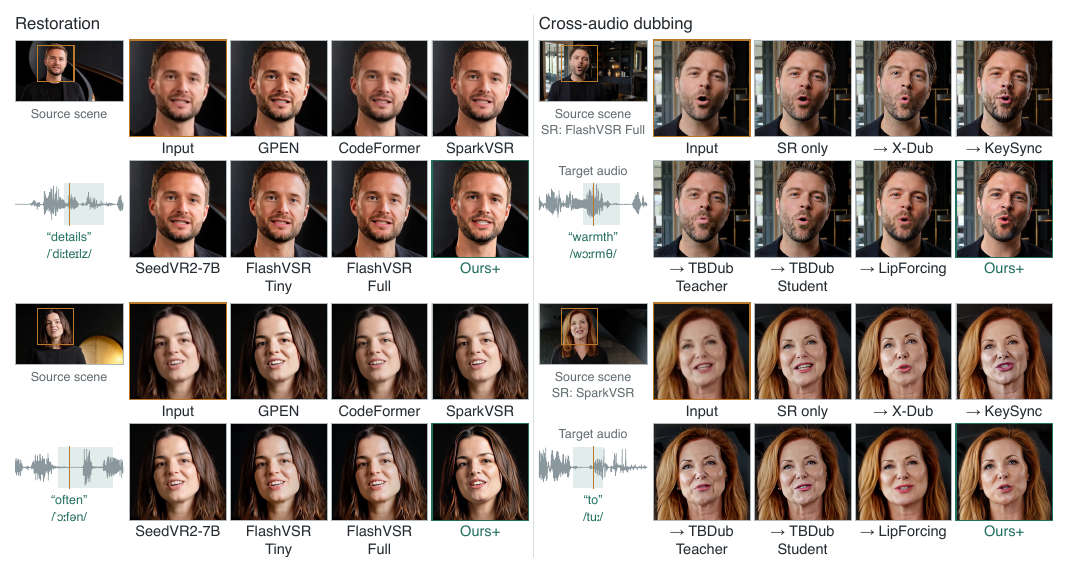}
  \caption{\textbf{Facial detail and articulation in AIGC scenes.} Four MiniMax-H3~\cite{minimax2026h3} clips are shown with matched-time head crops located by orange boxes. \textbf{Left:} restoration under original audio. \textbf{Right:} FlashVSR Full (top) or SparkVSR (bottom) is followed by the arrow-marked dubbing methods; Input and SR-only outputs provide visual controls. Ours$^{+}$ jointly reconstructs detail and articulation with a GPT-Image-enhanced reference, without an external SR frontend. Waveforms and word-level IPA labels identify the audio segments. Displayed predictions are unretouched.}
  \label{fig:qualitative}
\end{figure*}
Figure~\ref{fig:qualitative} compares matched-time facial crops from four native H3 clips under original and replacement audio.

Under original audio, the male example shows soft tooth boundaries in FlashVSR Tiny and Full and weak separation between adjacent teeth in SparkVSR. Ours$^{+}$ produces clearer upper-tooth boundaries and a more distinct lip--skin boundary, together with detailed beard texture. In the female example, it recovers distinct lip contours and visible upper teeth alongside fine hair detail.

Under replacement audio, the crops show differences in both the mouth and its surroundings. KeySync changes the beard and chin boundary after FlashVSR Full restoration, and alters lip geometry and color after SparkVSR restoration. LipForcing leaves the lower face comparatively soft in the displayed SparkVSR example. Ours$^{+}$ retains detailed beard texture around the male subject's mouth and produces a distinct lip contour with visible upper teeth for the female subject.

\subsection{Ablation Study}
\label{sec:ablations}
\begin{figure}[H]
  \centering
  \includegraphics[width=\textwidth]{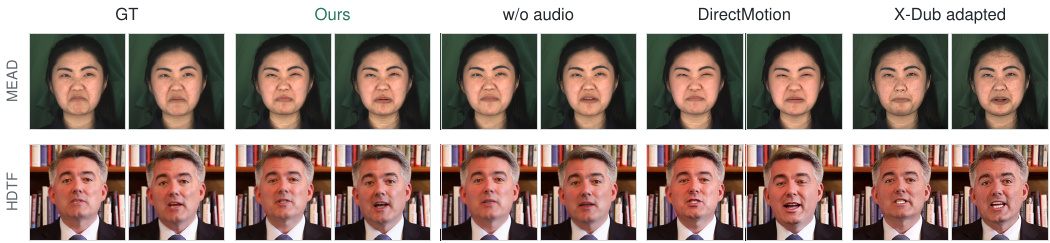}
  \caption{\textbf{Qualitative ablation comparison.} Each row shows two selected frames from one clip: MEAD~\cite{wang2020mead} (top) and HDTF~\cite{zhang2021hdtf} (bottom). The same frame pair is displayed for GT and each model variant, with ten images per row. All variants use the GT first frame, and predictions are extracted directly from native RGB outputs.}
  \label{fig:ablation-qualitative}
\end{figure}

We evaluate four configurations on the same 30 degraded videos using the unenhanced GT first frame. Original audio is supplied to all audio-conditioned variants. Alongside audio removal, restoration-adapted X-Dub and DirectMotion test alternative source-conditioning and history-conditioning configurations. Table~\ref{tab:ablations} reports quantitative results, and Figure~\ref{fig:ablation-qualitative} shows matched-frame comparisons.
\begin{table}[!htbp]
  \centering
  \caption{Ablation comparison on the same 30 videos: 15 from MEAD~\cite{wang2020mead} and 15 from HDTF~\cite{zhang2021hdtf}. All variants use the GT first frame. X-Dub~\cite{he2026xdub} and DirectMotion provide alternatives to source--noise mixing and structured reference injection, respectively. Values are video-wise means; GT is excluded from rankings. Dark and light green mark the best and second-best model results.}
  \label{tab:ablations}
  \setlength{\tabcolsep}{7pt}
  \renewcommand{\arraystretch}{1}
  \arrayrulecolor{black!30}
  \papertablefont
  \begin{tabular}{@{}lrrrrr@{}}
    \toprule
    Metric & \multicolumn{1}{c}{GT} & \multicolumn{1}{c}{Ours} & \multicolumn{1}{c}{\shortstack{w/o\\Audio}} & \multicolumn{1}{c}{\shortstack{Direct\\Motion}} & \multicolumn{1}{c}{\shortstack{X-Dub\\(adapted)}} \\
    \midrule
    PSNR $\uparrow$ & -- & \second{29.10} & \best{29.36} & 28.24 & 26.63 \\
    SSIM $\uparrow$ & -- & \second{0.8852} & \best{0.8889} & 0.8600 & 0.8415 \\
    LPIPS $\downarrow$ & -- & \best{0.0850} & \second{0.0856} & 0.1341 & 0.1523 \\
    \midrule
    MUSIQ $\uparrow$ & 63.26 & 68.19 & 69.09 & \second{69.49} & \best{70.08} \\
    CLIP-IQA $\uparrow$ & 0.4971 & 0.5591 & 0.5709 & \second{0.5807} & \best{0.6145} \\
    FasterVQA $\uparrow$ & 0.7880 & 0.8352 & \second{0.8432} & \best{0.8595} & 0.8354 \\
    DOVER $\uparrow$ & 0.6392 & 0.7367 & \second{0.7567} & \best{0.7602} & 0.7232 \\
    \midrule
    Sync-C $\uparrow$ & 7.762 & \best{7.621} & 4.283 & 7.254 & \second{7.407} \\
    Sync-D $\downarrow$ & 7.374 & \best{7.560} & 10.524 & \second{7.875} & 8.144 \\
    \bottomrule
  \end{tabular}
\end{table}

\paragraph{Source--noise reconstruction.}
Restoration-adapted X-Dub~\cite{he2026xdub} starts from Gaussian noise with dense degraded-video conditions. It uses our reference injection: an independently encoded GT first frame and motion history occupy separate slots. Ours improves PSNR from 26.63 to 29.10, reduces LPIPS from 0.1523 to 0.0850, and improves Sync-C/D from 7.407/8.144 to 7.621/7.560. Despite higher MUSIQ and CLIP-IQA, the adapted outputs show over-emphasized skin texture and hard dental edges in the selected frames. The comparison favors source--noise reconstruction for paired fidelity and synchronization.

\paragraph{Audio conditioning.}
Removing audio reduces Sync-C from 7.621 to 4.283 and increases Sync-D from 7.560 to 10.524, although PSNR improves by 0.26\,dB and LPIPS remains nearly unchanged at 0.0856 versus 0.0850. In the HDTF example, the visual-only variant narrows the target mouth opening, whereas Ours recovers it. Audio thus improves synchronization despite similar overall reconstruction fidelity.

\paragraph{Appearance and history conditioning.}
DirectMotion replaces the reference canvas with motion-frame conditioning: the first window repeats the GT reference five times, and later windows use the preceding output's last five frames without independently injecting the fixed reference. Ours reduces LPIPS from 0.1341 to 0.0850, improves PSNR from 28.24 to 29.10, and improves Sync-C/D from 7.254/7.875 to 7.621/7.560. DirectMotion's over-smoothed texture and artificial dot patterns in the selected outputs support retaining the fixed appearance reference alongside generated history.

\section{Conclusion}
TaoMate-FVSR combines facial restoration and lip correction in a single audio-guided model. It improves paired perceptual similarity and synchronization over the evaluated restoration baselines and performs favorably against restoration--dubbing cascades. Native AIGC examples show clearer facial detail and mouth boundaries. These results support speech-guided reconstruction when the source's visual evidence is unreliable.

Reconstructed detail is a generative estimate of appearance. Applications should obtain appropriate permission and disclose edited content to reduce impersonation risk.

\FloatBarrier
\bibliographystyle{unsrtnat}
\bibliography{references}

\begin{thebibliography}{41}
\providecommand{\natexlab}[1]{#1}
\providecommand{\url}[1]{\texttt{#1}}
\expandafter\ifx\csname urlstyle\endcsname\relax
  \providecommand{\doi}[1]{doi: #1}\else
  \providecommand{\doi}{doi: \begingroup \urlstyle{rm}\Url}\fi

\bibitem[{Wan Team} et~al.(2025)]{wan2025}
{Wan Team} et~al.
\newblock {Wan}: Open and advanced large-scale video generative models.
\newblock \emph{arXiv preprint arXiv:2503.20314}, 2025.
\newblock URL \url{https://arxiv.org/abs/2503.20314}.

\bibitem[Wang et~al.(2025)Wang, Lin, Lin, Ren, Wei, Yue, Zhou, Chen, Zhao, Yang, Xiao, Loy, and Jiang]{wang2025seedvr2}
Jianyi Wang, Shanchuan Lin, Zhijie Lin, Yuxi Ren, Meng Wei, Zongsheng Yue, Shangchen Zhou, Hao Chen, Yang Zhao, Ceyuan Yang, Xuefeng Xiao, Chen~Change Loy, and Lu~Jiang.
\newblock {SeedVR2}: One-step video restoration via diffusion adversarial post-training.
\newblock arXiv preprint arXiv:2506.05301v1, 2025.
\newblock URL \url{https://arxiv.org/abs/2506.05301v1}.

\bibitem[Zhuang et~al.(2025)Zhuang, Guo, Cai, Li, Liu, Yuan, and Xue]{zhuang2025flashvsr}
Junhao Zhuang, Shi Guo, Xin Cai, Xiaohui Li, Yihao Liu, Chun Yuan, and Tianfan Xue.
\newblock {FlashVSR}: Towards real-time diffusion-based streaming video super-resolution.
\newblock arXiv preprint arXiv:2510.12747v1, 2025.
\newblock URL \url{https://arxiv.org/abs/2510.12747v1}.

\bibitem[Yu et~al.(2026)Yu, Gao, Verlani, Gadde, Wang, Adsumilli, and Tu]{yu2026sparkvsr}
Jiongze Yu, Xiangbo Gao, Pooja Verlani, Akshay Gadde, Yilin Wang, Balu Adsumilli, and Zhengzhong Tu.
\newblock {SparkVSR}: Interactive video super-resolution via sparse keyframe propagation.
\newblock arXiv preprint arXiv:2603.16864, 2026.
\newblock URL \url{https://arxiv.org/abs/2603.16864}.

\bibitem[Hegde et~al.(2022)Hegde, Mukhopadhyay, Namboodiri, and Jawahar]{hegde2022extreme}
Sindhu~B. Hegde, Rudrabha Mukhopadhyay, Vinay~P. Namboodiri, and C.~V. Jawahar.
\newblock Extreme-scale talking-face video upsampling with audio-visual priors.
\newblock In \emph{Proceedings of the 30th ACM International Conference on Multimedia}, 2022.
\newblock \doi{10.1145/3503161.3548080}.
\newblock URL \url{https://doi.org/10.1145/3503161.3548080}.

\bibitem[Zhang and Wu(2023)]{zhang2023mdvd}
Xi~Zhang and Xiaolin Wu.
\newblock Multi-modality deep restoration of extremely compressed face videos.
\newblock \emph{IEEE Transactions on Pattern Analysis and Machine Intelligence}, 45\penalty0 (2):\penalty0 2024--2037, 2023.
\newblock \doi{10.1109/TPAMI.2022.3157388}.
\newblock URL \url{https://doi.org/10.1109/TPAMI.2022.3157388}.

\bibitem[Cao et~al.(2026)Cao, Gao, Sun, Liu, Zhang, and Min]{cao2026gavn}
Yuqin Cao, Yixuan Gao, Wei Sun, Xiaohong Liu, Yulun Zhang, and Xiongkuo Min.
\newblock Audio-assisted face video restoration with temporal and identity complementary learning.
\newblock In \emph{Proceedings of the AAAI Conference on Artificial Intelligence}, volume~40, pages 2661--2669, 2026.
\newblock \doi{10.1609/aaai.v40i4.37254}.
\newblock URL \url{https://ojs.aaai.org/index.php/AAAI/article/view/37254}.

\bibitem[He et~al.(2026)He, Zhang, Chen, Zheng, Chen, Tang, Huang, Liu, Wan, and Wu]{he2026xdub}
Xu~He, Haoxian Zhang, Hejia Chen, Changyuan Zheng, Liyang Chen, Songlin Tang, Jiehui Huang, Xiaoqiang Liu, Pengfei Wan, and Zhiyong Wu.
\newblock From inpainting to editing: Unlocking robust mask-free visual dubbing via generative bootstrapping.
\newblock arXiv preprint arXiv:2512.25066v2, 2026.
\newblock URL \url{https://arxiv.org/abs/2512.25066v2}.

\bibitem[Cheng et~al.(2022)Cheng, Cun, Zhang, Xia, Yin, Zhu, Wang, Wang, and Wang]{cheng2022videoretalking}
Kun Cheng, Xiaodong Cun, Yong Zhang, Menghan Xia, Fei Yin, Mingrui Zhu, Xuan Wang, Jue Wang, and Nannan Wang.
\newblock {VideoReTalking}: Audio-based lip synchronization for talking head video editing in the wild.
\newblock arXiv preprint arXiv:2211.14758, 2022.
\newblock URL \url{https://arxiv.org/abs/2211.14758}.

\bibitem[Wang et~al.(2021)Wang, Li, Zhang, and Shan]{wang2021gfpgan}
Xintao Wang, Yu~Li, Honglun Zhang, and Ying Shan.
\newblock Towards real-world blind face restoration with generative facial prior.
\newblock In \emph{Proceedings of the IEEE/CVF Conference on Computer Vision and Pattern Recognition}, 2021.
\newblock URL \url{https://openaccess.thecvf.com/content/CVPR2021/papers/Wang_Towards_Real-World_Blind_Face_Restoration_With_Generative_Facial_Prior_CVPR_2021_paper.pdf}.

\bibitem[Yang et~al.(2021)Yang, Ren, Xie, and Zhang]{yang2021gpen}
Tao Yang, Peiran Ren, Xuansong Xie, and Lei Zhang.
\newblock {GAN} prior embedded network for blind face restoration in the wild.
\newblock In \emph{Proceedings of the IEEE/CVF Conference on Computer Vision and Pattern Recognition}, pages 672--681, 2021.
\newblock URL \url{https://openaccess.thecvf.com/content/CVPR2021/html/Yang_GAN_Prior_Embedded_Network_for_Blind_Face_Restoration_in_the_CVPR_2021_paper.html}.

\bibitem[Zhou et~al.(2022)Zhou, Chan, Li, and Loy]{zhou2022codeformer}
Shangchen Zhou, Kelvin Chan, Chongyi Li, and Chen~Change Loy.
\newblock Towards robust blind face restoration with codebook lookup transformer.
\newblock In \emph{Advances in Neural Information Processing Systems}, volume~35, pages 30599--30611, 2022.
\newblock URL \url{https://proceedings.neurips.cc/paper_files/paper/2022/hash/c573258c38d0a3919d8c1364053c45df-Abstract-Conference.html}.

\bibitem[Wang et~al.(2024)Wang, Yue, Zhou, Chan, and Loy]{wang2024stablesr}
Jianyi Wang, Zongsheng Yue, Shangchen Zhou, Kelvin C.~K. Chan, and Chen~Change Loy.
\newblock Exploiting diffusion prior for real-world image super-resolution.
\newblock \emph{International Journal of Computer Vision}, 2024.
\newblock URL \url{https://github.com/IceClear/StableSR}.

\bibitem[Lin et~al.(2024)Lin, He, Chen, Lyu, Dai, Yu, Qiao, Ouyang, and Dong]{lin2024diffbir}
Xinqi Lin, Jingwen He, Ziyan Chen, Zhaoyang Lyu, Bo~Dai, Fanghua Yu, Yu~Qiao, Wanli Ouyang, and Chao Dong.
\newblock {DiffBIR}: Toward blind image restoration with generative diffusion prior.
\newblock In \emph{European Conference on Computer Vision}, 2024.
\newblock URL \url{https://www.ecva.net/papers/eccv_2024/papers_ECCV/papers/07690.pdf}.

\bibitem[Yu et~al.(2024)Yu, Gu, Li, Hu, Kong, Wang, He, Qiao, and Dong]{yu2024supir}
Fanghua Yu, Jinjin Gu, Zheyuan Li, Jinfan Hu, Xiangtao Kong, Xintao Wang, Jingwen He, Yu~Qiao, and Chao Dong.
\newblock Scaling up to excellence: Practicing model scaling for photo-realistic image restoration in the wild.
\newblock In \emph{Proceedings of the IEEE/CVF Conference on Computer Vision and Pattern Recognition}, pages 25669--25680, 2024.
\newblock URL \url{https://openaccess.thecvf.com/content/CVPR2024/html/Yu_Scaling_Up_to_Excellence_Practicing_Model_Scaling_for_Photo-Realistic_Image_CVPR_2024_paper.html}.

\bibitem[Chan et~al.(2022)Chan, Zhou, Xu, and Loy]{chan2022basicvsrpp}
Kelvin C.~K. Chan, Shangchen Zhou, Xiangyu Xu, and Chen~Change Loy.
\newblock {BasicVSR++}: Improving video super-resolution with enhanced propagation and alignment.
\newblock In \emph{Proceedings of the IEEE/CVF Conference on Computer Vision and Pattern Recognition}, pages 5972--5981, 2022.
\newblock URL \url{https://openaccess.thecvf.com/content/CVPR2022/html/Chan_BasicVSR_Improving_Video_Super-Resolution_With_Enhanced_Propagation_and_Alignment_CVPR_2022_paper.html}.

\bibitem[Feng et~al.(2024)Feng, Li, and Loy]{feng2024keep}
Ruicheng Feng, Chongyi Li, and Chen~Change Loy.
\newblock Kalman-inspired {FEaturE} propagation for video face super-resolution.
\newblock In \emph{European Conference on Computer Vision}, 2024.
\newblock URL \url{https://jnjaby.github.io/projects/KEEP/}.

\bibitem[Zhou et~al.(2024)Zhou, Yang, Wang, Luo, and Loy]{zhou2024upscaleavideo}
Shangchen Zhou, Peiqing Yang, Jianyi Wang, Yihang Luo, and Chen~Change Loy.
\newblock {Upscale-A-Video}: Temporal-consistent diffusion model for real-world video super-resolution.
\newblock In \emph{Proceedings of the IEEE/CVF Conference on Computer Vision and Pattern Recognition}, 2024.
\newblock URL \url{https://openaccess.thecvf.com/content/CVPR2024/papers/Zhou_Upscale-A-Video_Temporal-Consistent_Diffusion_Model_for_Real-World_Video_Super-Resolution_CVPR_2024_paper.pdf}.

\bibitem[Zhang et~al.(2025)Zhang, Alaluf, Ma, Kadambi, Wang, and Aberman]{zhang2025instantrestore}
Howard Zhang, Yuval Alaluf, Sizhuo Ma, Achuta Kadambi, Jian Wang, and Kfir Aberman.
\newblock {InstantRestore}: Single-step personalized face restoration with shared-image attention.
\newblock Official SIGGRAPH implementation, 2025.
\newblock URL \url{https://github.com/snap-research/InstantRestore}.

\bibitem[Eteke et~al.(2026)Eteke, Tosun, and Steinbach]{eteke2026rgfvrpaper}
Cem Eteke, Batuhan Tosun, and Eckehard Steinbach.
\newblock {RGFVR}: Reference-guided face video restoration with flow matching, 2026.
\newblock URL \url{https://arxiv.org/abs/2606.16401}.

\bibitem[Li et~al.(2023)Li, Qin, Liang, and Wei]{li2023hdtr}
Yongyuan Li, Xiuyuan Qin, Chao Liang, and Mingqiang Wei.
\newblock {HDTR-Net}: A real-time high-definition teeth restoration network for arbitrary talking face generation methods.
\newblock In \emph{Chinese Conference on Pattern Recognition and Computer Vision (PRCV)}, pages 89--103. Springer, 2023.
\newblock URL \url{https://arxiv.org/abs/2309.07495}.

\bibitem[Su et~al.(2025)Su, Zhang, Wei, Yang, and Deng]{su2025orthonet}
Zibo Su, Ziqi Zhang, Kun Wei, Xu~Yang, and Cheng Deng.
\newblock Outstanding orthodontist: No more artifactual teeth in talking face.
\newblock In \emph{Proceedings of the Thirty-Fourth International Joint Conference on Artificial Intelligence}, pages 1891--1899, 2025.
\newblock \doi{10.24963/ijcai.2025/211}.
\newblock URL \url{https://ijcai.org/proceedings/2025/211}.

\bibitem[Prajwal et~al.(2020)Prajwal, Mukhopadhyay, Namboodiri, and Jawahar]{prajwal2020wav2lip}
K~R Prajwal, Rudrabha Mukhopadhyay, Vinay~P. Namboodiri, and C.~V. Jawahar.
\newblock A lip sync expert is all you need for speech to lip generation in the wild.
\newblock In \emph{Proceedings of the 28th ACM International Conference on Multimedia}, pages 484--492, 2020.
\newblock \doi{10.1145/3394171.3413532}.
\newblock URL \url{https://doi.org/10.1145/3394171.3413532}.

\bibitem[Bigata et~al.(2025)Bigata, Mira, Bounareli, Stypu{\l}kowski, Vougioukas, Petridis, and Pantic]{bigata2025keysync}
Antoni Bigata, Rodrigo Mira, Stella Bounareli, Micha{\l} Stypu{\l}kowski, Konstantinos Vougioukas, Stavros Petridis, and Maja Pantic.
\newblock {KeySync}: A robust approach for leakage-free lip synchronization in high resolution.
\newblock arXiv preprint arXiv:2505.00497, 2025.
\newblock URL \url{https://arxiv.org/abs/2505.00497}.

\bibitem[Li et~al.(2026)Li, Li, Xu, Jin, and Ma]{li2026tbdub}
Bihan Li, Xinyang Li, Zeran Xu, Meiguang Jin, and Junfeng Ma.
\newblock {TBDub}: Production-oriented visual dubbing, 2026.
\newblock URL \url{https://arxiv.org/abs/2609.06144}.

\bibitem[Cho et~al.(2026)Cho, Jang, Lee, Lee, Jin, Shin, Yi, Park, Park, and Kim]{cho2026lipforcing}
Paul~Hyunbin Cho, Jinhyuk Jang, SeokYoung Lee, Joungbin Lee, Siyoon Jin, Heeseong Shin, Jung Yi, Yunjin Park, Chulmin Park, and Seungryong Kim.
\newblock Lip forcing: Few-step autoregressive diffusion for real-time lip synchronization.
\newblock arXiv preprint arXiv:2606.11180, 2026.
\newblock URL \url{https://arxiv.org/abs/2606.11180}.

\bibitem[Lipman et~al.(2022)Lipman, Chen, Ben-Hamu, Nickel, and Le]{lipman2022flowmatching}
Yaron Lipman, Ricky T.~Q. Chen, Heli Ben-Hamu, Maximilian Nickel, and Matt Le.
\newblock Flow matching for generative modeling.
\newblock arXiv preprint arXiv:2210.02747v1, 2022.
\newblock URL \url{https://arxiv.org/abs/2210.02747v1}.

\bibitem[Hsu et~al.(2021)Hsu, Bolte, Tsai, Lakhotia, Salakhutdinov, and Mohamed]{hsu2021hubert}
Wei-Ning Hsu, Benjamin Bolte, Yao-Hung~Hubert Tsai, Kushal Lakhotia, Ruslan Salakhutdinov, and Abdelrahman Mohamed.
\newblock {HuBERT}: Self-supervised speech representation learning by masked prediction of hidden units.
\newblock \emph{IEEE/ACM Transactions on Audio, Speech, and Language Processing}, 29:\penalty0 3451--3460, 2021.
\newblock \doi{10.1109/TASLP.2021.3122291}.
\newblock URL \url{https://doi.org/10.1109/TASLP.2021.3122291}.

\bibitem[Zhang et~al.(2018)Zhang, Isola, Efros, Shechtman, and Wang]{zhang2018lpips}
Richard Zhang, Phillip Isola, Alexei~A. Efros, Eli Shechtman, and Oliver Wang.
\newblock The unreasonable effectiveness of deep features as a perceptual metric.
\newblock In \emph{Proceedings of the IEEE Conference on Computer Vision and Pattern Recognition}, pages 586--595, 2018.
\newblock URL \url{https://openaccess.thecvf.com/content_cvpr_2018/html/Zhang_The_Unreasonable_Effectiveness_CVPR_2018_paper.html}.

\bibitem[Wang and Bovik(2009)]{wang2009mse}
Zhou Wang and Alan~C. Bovik.
\newblock Mean squared error: Love it or leave it? a new look at signal fidelity measures.
\newblock \emph{IEEE Signal Processing Magazine}, 26\penalty0 (1):\penalty0 98--117, 2009.
\newblock \doi{10.1109/MSP.2008.930649}.
\newblock URL \url{https://doi.org/10.1109/MSP.2008.930649}.

\bibitem[Wang et~al.(2004)Wang, Bovik, Sheikh, and Simoncelli]{wang2004ssim}
Zhou Wang, Alan~C. Bovik, Hamid~R. Sheikh, and Eero~P. Simoncelli.
\newblock Image quality assessment: From error visibility to structural similarity.
\newblock \emph{IEEE Transactions on Image Processing}, 13\penalty0 (4):\penalty0 600--612, 2004.
\newblock \doi{10.1109/TIP.2003.819861}.
\newblock URL \url{https://doi.org/10.1109/TIP.2003.819861}.

\bibitem[Ding et~al.(2022)Ding, Ma, Wang, and Simoncelli]{ding2022dists}
Keyan Ding, Kede Ma, Shiqi Wang, and Eero~P. Simoncelli.
\newblock Image quality assessment: Unifying structure and texture similarity.
\newblock \emph{IEEE Transactions on Pattern Analysis and Machine Intelligence}, 44\penalty0 (5):\penalty0 2567--2581, 2022.
\newblock \doi{10.1109/TPAMI.2020.3045810}.
\newblock URL \url{https://doi.org/10.1109/TPAMI.2020.3045810}.

\bibitem[Chung and Zisserman(2016)]{chung2016syncnet}
Joon~Son Chung and Andrew Zisserman.
\newblock Out of time: Automated lip sync in the wild.
\newblock In \emph{Workshop on Multi-view Lip-reading, ACCV}, 2016.
\newblock URL \url{https://www.robots.ox.ac.uk/~vgg/publications/2016/Chung16a/}.

\bibitem[Ke et~al.(2021)Ke, Wang, Wang, Milanfar, and Yang]{ke2021musiq}
Junjie Ke, Qifei Wang, Yilin Wang, Peyman Milanfar, and Feng Yang.
\newblock {MUSIQ}: Multi-scale image quality transformer.
\newblock In \emph{Proceedings of the IEEE/CVF International Conference on Computer Vision}, pages 5148--5157, 2021.
\newblock URL \url{https://openaccess.thecvf.com/content/ICCV2021/html/Ke_MUSIQ_Multi-Scale_Image_Quality_Transformer_ICCV_2021_paper.html}.

\bibitem[Wang et~al.(2023)Wang, Chan, and Loy]{wang2023clipiqa}
Jianyi Wang, Kelvin C.~K. Chan, and Chen~Change Loy.
\newblock Exploring {CLIP} for assessing the look and feel of images.
\newblock In \emph{Proceedings of the AAAI Conference on Artificial Intelligence}, volume~37, pages 2555--2563, 2023.
\newblock \doi{10.1609/aaai.v37i2.25353}.
\newblock URL \url{https://doi.org/10.1609/aaai.v37i2.25353}.

\bibitem[Wu et~al.(2023{\natexlab{a}})Wu, Chen, Liao, Hou, Sun, Yan, Gu, and Lin]{wu2022fasterquality}
Haoning Wu, Chaofeng Chen, Liang Liao, Jingwen Hou, Wenxiu Sun, Qiong Yan, Jinwei Gu, and Weisi Lin.
\newblock Neighbourhood representative sampling for efficient end-to-end video quality assessment.
\newblock \emph{IEEE Transactions on Pattern Analysis and Machine Intelligence}, 45\penalty0 (12):\penalty0 15185--15202, 2023{\natexlab{a}}.
\newblock \doi{10.1109/TPAMI.2023.3319332}.
\newblock URL \url{https://doi.org/10.1109/TPAMI.2023.3319332}.

\bibitem[Wu et~al.(2023{\natexlab{b}})Wu, Zhang, Liao, Chen, Hou, Wang, Sun, Yan, and Lin]{wu2023dover}
Haoning Wu, Erli Zhang, Liang Liao, Chaofeng Chen, Jingwen Hou, Annan Wang, Wenxiu Sun, Qiong Yan, and Weisi Lin.
\newblock Exploring video quality assessment on user generated contents from aesthetic and technical perspectives.
\newblock In \emph{Proceedings of the IEEE/CVF International Conference on Computer Vision}, pages 20144--20154, 2023{\natexlab{b}}.
\newblock URL \url{https://openaccess.thecvf.com/content/ICCV2023/html/Wu_Exploring_Video_Quality_Assessment_on_User_Generated_Contents_from_Aesthetic_ICCV_2023_paper.html}.

\bibitem[Wang et~al.(2020)Wang, Wu, Song, Yang, Wu, Qian, He, Qiao, and Loy]{wang2020mead}
Kaisiyuan Wang, Qianyi Wu, Linsen Song, Zhuoqian Yang, Wayne Wu, Chen Qian, Ran He, Yu~Qiao, and Chen~Change Loy.
\newblock {MEAD}: A large-scale audio-visual dataset for emotional talking-face generation.
\newblock In \emph{Proceedings of the European Conference on Computer Vision}, 2020.
\newblock URL \url{https://wywu.github.io/projects/MEAD/MEAD.html}.

\bibitem[Zhang et~al.(2021)Zhang, Li, Ding, and Fan]{zhang2021hdtf}
Zhimeng Zhang, Lincheng Li, Yu~Ding, and Changjie Fan.
\newblock Flow-guided one-shot talking face generation with a high-resolution audio-visual dataset.
\newblock In \emph{Proceedings of the IEEE/CVF Conference on Computer Vision and Pattern Recognition}, pages 3661--3670, 2021.
\newblock URL \url{https://openaccess.thecvf.com/content/CVPR2021/html/Zhang_Flow-Guided_One-Shot_Talking_Face_Generation_With_a_High-Resolution_Audio-Visual_Dataset_CVPR_2021_paper.html}.

\bibitem[Chen et~al.(2025)Chen, Huang, Liu, Ye, Chen, Zhu, Guan, Wang, Chen, Li, Lim, Yang, and Wang]{chen2025talkvid}
Shunian Chen, Hejin Huang, Yexin Liu, Zihan Ye, Pengcheng Chen, Chenghao Zhu, Michael Guan, Rongsheng Wang, Junying Chen, Guanbin Li, Ser-Nam Lim, Harry Yang, and Benyou Wang.
\newblock {TalkVid}: A large-scale diversified dataset for audio-driven talking head synthesis.
\newblock arXiv preprint arXiv:2508.13618, 2025.
\newblock URL \url{https://arxiv.org/abs/2508.13618}.

\bibitem[{MiniMax}(2026)]{minimax2026h3}
{MiniMax}.
\newblock {MiniMax H3}: An open model breaking the boundaries between tasks and modalities.
\newblock MiniMax Research, July 2026.
\newblock URL \url{https://www.minimax.io/blog/minimax-h3}.
\newblock \url{https://www.minimax.io/blog/minimax-h3}.

\end{thebibliography}
\end{document}